\documentclass{article}

\PassOptionsToPackage{numbers, compress}{natbib}

\usepackage[preprint]{neurips_2024}

\usepackage[utf8]{inputenc} 
\usepackage[T1]{fontenc}    
\usepackage{hyperref}       
\usepackage{url}            
\usepackage{booktabs}       
\usepackage{amsfonts}       
\usepackage{nicefrac}       
\usepackage{microtype}      
\usepackage{xcolor}         
\usepackage{graphicx}
\usepackage{color}
\usepackage{multirow} 
\usepackage{wrapfig}
\newcommand\mamba{\raisebox{-5pt}{\includegraphics[width=1.5em]{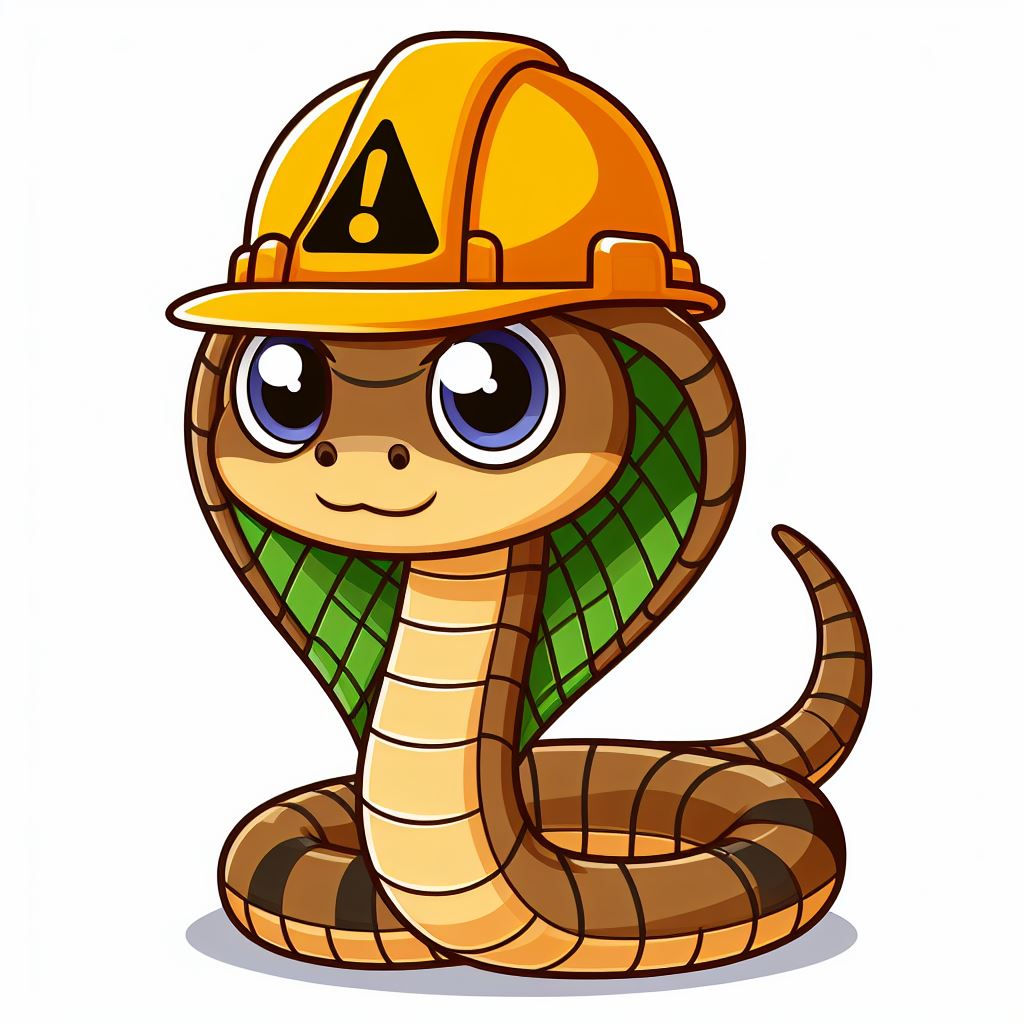}}}

\usepackage{amsmath,amsfonts,bm}

\def\eqref#1{equation~\ref{#1}}

\def\1{\bm{1}}

\def\vh{{\bm{h}}}

\def\vp{{\bm{p}}}

\def\vv{{\bm{v}}}
\def\vw{{\bm{w}}}

\def\vy{{\bm{y}}}
\def\vz{{\bm{z}}}

\def\mF{{\bm{F}}}

\def\mI{{\bm{I}}}

\def\mM{{\bm{M}}}
\def\mN{{\bm{N}}}

\def\mS{{\bm{S}}}

\def\mZ{{\bm{Z}}}

\DeclareMathAlphabet{\mathsfit}{\encodingdefault}{\sfdefault}{m}{sl}
\SetMathAlphabet{\mathsfit}{bold}{\encodingdefault}{\sfdefault}{bx}{n}

\def\gL{{\mathcal{L}}}

\newcommand{\R}{\mathbb{R}}

\newcommand{\softmax}{\mathrm{softmax}}

\DeclareMathOperator*{\MLP}{MLP}
\DeclareMathOperator*{\PatchEmbed}{PatchEmbed}
\DeclareMathOperator*{\MemSSMBlock}{Mem-SSM-Block}
\DeclareMathOperator*{\Linear}{Linear}
\DeclareMathOperator*{\MemoryEncode}{MemoryEncode}
\DeclareMathOperator*{\FusionModule}{FusionModule}
\DeclareMathOperator*{\Fusion}{Fusion}
\DeclareMathOperator*{\SelectiveScan}{SelectiveScan}
\DeclareMathOperator*{\LayerNorm}{LayerNorm}
\DeclareMathOperator*{\Convolution}{Conv}
\DeclareMathOperator*{\Pooling}{Pooling}
\DeclareMathOperator*{\similarity}{Similarity}
\DeclareMathOperator*{\contrastive}{Contrastive}

\title{\mamba MemoryMamba: Memory-Augmented State Space Model for Defect Recognition}

\author{%
Qianning Wang$^{1}$, He Hu$^{2}$, Yucheng Zhou$^{3}$ \\
  $^{1}$Nanjing Audit University, $^{2}$Xi’an University of Science and Technology\\
  $^{3}$SKL-IOTSC, CIS, University of Macau \\
  \texttt{yucheng.zhou@connect.um.edu.mo} \\
}

\begin{document}

\maketitle

\begin{abstract}
As automation advances in manufacturing, the demand for precise and sophisticated defect detection technologies grows. Existing vision models for defect recognition methods are insufficient for handling the complexities and variations of defects in contemporary manufacturing settings. These models especially struggle in scenarios involving limited or imbalanced defect data. In this work, we introduce MemoryMamba, a novel memory-augmented state space model (SSM), designed to overcome the limitations of existing defect recognition models. MemoryMamba integrates the state space model with the memory augmentation mechanism, enabling the system to maintain and retrieve essential defect-specific information in training. Its architecture is designed to capture dependencies and intricate defect characteristics, which are crucial for effective defect detection. In the experiments, MemoryMamba was evaluated across four industrial datasets with diverse defect types and complexities. The model consistently outperformed other methods, demonstrating its capability to adapt to various defect recognition scenarios.
\end{abstract}

\section{Introduction}
The advent of deep learning technologies has significantly advanced various industries~\cite{chen2023point,liu2024news,lai2024language,chen2023class,chen2021multimodal}, particularly manufacturing, by transforming defect recognition processes essential for quality control \cite{ren2022state,su2024large}. In manufacturing, defect recognition plays a pivotal role in enhancing production efficiency, reducing costs, and ensuring product reliability. The demand for sophisticated and precise defect recognition systems has intensified as industries progress towards more automated and precise manufacturing techniques. These systems are essential not only for reducing the incidence of defective products reaching consumers but also for enhancing the overall sustainability of production lines through waste reduction and improved safety protocols.

The advent of Convolutional Neural Networks (CNNs) has significantly transformed defect recognition, enhancing both accuracy and automation \cite{wang2018fast,zhang2017application}. These models span from adaptations of established architectures like VGG \cite{apostolopoulos2020industrial} and ResNet \cite{Rafiei_2023} to more intricate configurations such as  T-CNN \cite{martinramiro2023boosting} and GCNN \cite{wang2022graph}, have significantly elevated the precision and speed of defect recognition systems \cite{jha2023deep}. The incorporation of techniques such as transfer learning and classifier fusion further enriches their adaptability and robustness across diverse manufacturing environments. Nevertheless, these models frequently encounter challenges in scenarios characterized by limited or highly imbalanced defect samples, i.e., conditions that are prevalent in specialized industrial settings. The dependency of these models on extensively annotated datasets to achieve high performance is a significant limitation, particularly in environments where such data is scarce.

State Space Models (SSMs) \cite{hamilton1994state} have recently provided new avenues to address these challenges effectively. The Mamba model \cite{gu2023mamba} and its variants in Computer Vision, such as VMamba \cite{liu2024vmamba} and VIM \cite{zhu2024vision}, have demonstrated considerable potential in improving vision recognition. VMamba reduces computational complexity by leveraging the Cross-Scan Module (CSM), which performs 1D selective scanning in a 2D image space, thereby enabling global receptive fields without the heavy computational cost associated with Vision Transformers (ViTs). On the other hand, VIM employs bidirectional state space models along with position embeddings to handle image sequences, which helps in capturing comprehensive visual data necessary for identifying subtle and complex defects. Moreover, the efficiency and scalability of the Mamba model, due to its hardware-aware design, make it ideal for deployment in industrial settings where real-time processing is essential. 

To address the above problems, we propose MemoryMamba, a novel memory-augmented state space model specifically designed for defect recognition tasks. The architecture of MemoryMamba combines state space techniques with memory augmentation to effectively capture dependencies and intricate defect characteristics. The model incorporates coarse- and fine-grained memory networks to better retain and access critical defect information from previously trained samples. Additionally, we introduce a fusion module that integrates features extracted from these memory networks, enhancing the model's capability. Optimization strategies based on contrastive learning and mutual information maximization are also proposed for coarse- and fine-grained memory networks, respectively. 

In the experiments, we evaluate the effectiveness of MemoryMamba by conducting comprehensive experiments across four industrial defect recognition datasets, encompassing a range of defect types and complexities. In addition, MemoryMamba consistently outperforms the existing models. 

The main contributions of our paper are as follows:
\begin{itemize}
\item To the best of our knowledge, MemoryMamba is the first state space model for defect recognition in industrial applications. 
\item The MemoryMamba model incorporates a novel memory-augmented mechanism that allows for the retention and efficient retrieval of critical defect-related information from historical data. 
\item We design the optimization methods for coarse- and fine-grained memory networks, and propose a fusion module to integrate the visual feature and memory vector. Moreover, we propose optimization strategies, based on contrastive learning and mutual information maximization, for coarse-grained and fine-grained memory networks, respectively. 
\item Our experiments conducted a comprehensive comparative analysis with existing defect detection models, demonstrating MemoryMamba's superior performance.
\end{itemize}

\section{Related Work}
\subsection{Defect Recognition}
With the advancement of deep learning applications across various industries~\cite{chen2024bridging,liu2024particle,zhou2024fine,zhou2023towards}, defect detection technologies have become crucial in enhancing product quality and operational efficiency within the manufacturing industry, particularly with the emergence of Industry 4.0. These technologies integrate machine learning and computer vision, transforming traditional defect detection methods to achieve unprecedented accuracy and efficiency \cite{bhatt2021image,cao2024survey,czimmermann2020visual}. Significant advancements include the Tensor Convolutional Neural Network (T-CNN) developed by\citet{martinramiro2023boosting}, which offers reduced parameter counts and improved training speeds without sacrificing accuracy. Similarly, \citet{shi2023center} introduced the Center-based Transfer Feature Learning with Classifier Adaptation (CTFLCA), effectively adapting to diverse image distributions in different manufacturing settings and achieving high defect detection accuracies. \citet{zaghdoudi2024steel} devise a classifier fusion approach for steel defect classification, combining SVM and RF with Bayesian rule, enhancing accuracy and speed. Moreover, there is an attempt to enhance the VGG19 into a Multipath VGG19, which improved defect detection across various datasets \citet{apostolopoulos2020industrial}. In contrast, \citet{fu2019recognition} effectively used a pretrained VGG16 model with a custom CNN classifier for detecting surface defects on steel strips, particularly in data-limited scenarios. 
In the area of X-ray imaging for quality control, \citet{Rafiei_2023} utilized a ResNet-based model optimized through structural pruning to enhance defect detection in mineral wool production. \citet{garcia2022automated} developed Automated Defect Recognition (ADR) systems using CNNs that boost the reliability and speed of industrial X-ray analysis while minimizing subjective discrepancies. \citet{gamdha2021automated} develop Sim-ADR, using synthetic data and ray tracing to train CNNs for X-ray anomaly detection with 87\% accuracy.
In addition, \citet{gao2019multilevel} proposed a method based on multilevel information fusion, ideal for small sample sizes, using a Gaussian pyramid with three VGG16 networks to enhance recognition accuracy. a subsequent work \cite{gao2020semi} involved a semi-supervised learning approach with CNNs enhanced by Pseudo-Label, leveraging both labeled and unlabeled data to improve defect detection. 

Different from CNN-based methods, \citet{wang2022graph} introduced the Graph guided Convolutional Neural Network (GCNN), which incorporates a graph guidance mechanism into CNNs to enhance feature extraction and manage variations within defect classes. a subsequent work \cite{wang2020deformable} introduced a deformable convolutional network (DC-Net) designed for mixed-type defect detection in wafer maps, achieving high accuracy with its novel structural design. \citet{yu2015wafer} present a manifold learning system with JLNDA for defect detection in wafer maps, outperforming traditional methods using WM-811K data. Specialized CNN adaptations for specific industrial applications include \citet{tao2018wire} for spring-wire sockets and Mentouri et al.\cite{mentouri2020improved} for online surface defect monitoring in the steel-making industry. \citet{yang2021image} developed a model for detecting wind turbine blade damage, demonstrating superior performance in challenging environments. Other works in defect detection include the use of advanced neural network models by \citet{konovalenko2020steel} and \citet{cheng2020retinanet} for rolled metal and steel surface defects. \citet{su2020deep} developed a Complementary Attention Network (CAN) within a faster R-CNN framework to refine defect detection in solar cell images. \citet{zhang2023improved} enhance DETR with ResNet, ECA-Net, dynamic anchors, and deformable attention for superior casting defect detection.

\subsection{State Space Models}
Different from Transformers~\cite{vaswani2017attention,zhou2023multimodal,zhou2023style,lai2024adaptive}, State Space Models (SSMs) continue to evolve, increasingly shaping the frontier of sequence modeling by tackling the challenges of computational efficiency and long-range dependency management across various data types \cite{wang2024state,gu2021efficiently,zhang2024survey}. For time series forecasting, \citet{xu2024integrating} introduced Mambaformer, a hybrid model combining Mamba and Transformer, efficiently managing both long and short-range dependencies and outperforming traditional models. \citet{liang2024bimamba4ts} propose Bi-Mamba4TS, a model enhancing long-term forecasting with efficient computation and adaptive tokenization, outperforming current methods in accuracy. 
In computer vision, \citet{chen2024mambauiesr} presented MambaUIE, optimizing state space models for Underwater Image Enhancement by efficiently combining global and local features, drastically reducing computational demands while maintaining high accuracy. 

For image restoration, \citet{deng2024cumamba} introduced the Channel-Aware U-Shaped Mamba (CU-Mamba) model, which incorporates a dual State Space Model into the U-Net architecture to efficiently encode global context and preserve channel correlations. 
In hyperspectral image denoising, \citet{liu2024hsidmamba} presented HSIDMamba, a Selective State Space Model that integrates advanced spatial-spectral mechanisms, significantly improving efficiency and performance, outperforming existing transformer-based methods by 30\%. 
Moreover, \citet{wang2024insectmamba} introduced InsectMamba, integrating State Space Models, CNNs, Multi-Head Self-Attention, and MLPs in Mix-SSM blocks to effectively extract detailed features for precise insect pest classification, demonstrating superior performance on various datasets. In the 3D point cloud analysis, \citet{han2024mamba3d} developed Mamba3D, which leverages Local Norm Pooling and a bidirectional SSM, significantly surpassing Transformer models in both accuracy and scalability. 

For medical imaging, several models demonstrate the application of SSMs: \citet{wu2024hvmunet} proposed H-vmunet, enhancing feature extraction with High-order 2D-selective-scan (H-SS2D) and Local-SS2D modules. \citet{yue2024medmamba} introduced MedMamba, leveraging Conv-SSM modules to efficiently capture long-range dependencies. \citet{ruan2024vmunet} developed VM-UNet, using Visual State Space blocks for enhanced contextual information capture. Furthermore, \citet{wu2024ultralight} presented UltraLight VM-UNet, a highly efficient model using a novel PVM Layer for parallel feature processing.

In enhancing the capabilities of SSMs, \citet{he2024densemamba} introduced DenseSSM, integrating shallow-layer hidden states into deeper layers to improve performance while maintaining efficiency. \citet{smith2023convolutional} introduced ConvS5, a convolutional state space model that excels in long spatiotemporal sequence modeling, training faster and generating samples more efficiently than competitors. \citet{fathullah2023multihead} developed MH-SSM, a multi-head state space model that surpasses the transformer transducer on LibriSpeech and achieves state-of-the-art results when integrated into the Stateformer.

\section{Preliminaries}
\subsection{State Space Models}
State space models (SSMs) constitute a robust framework for the analysis of time series data, encapsulating the dynamics of systems through a series of mathematical representations. These models articulate the time series as a function of latent states and observations, with the state equations delineating the evolution of these latent states, and the observation equations describing the measurements derived from these states.

The evolution of the state vector $\mathbf{x}_t$ at time $t$ is governed by the state transition equation:
\begin{align}
\mathbf{x}_t = \mathbf{F}_t \mathbf{x}_{t-1} + \mathbf{G}_t \mathbf{w}_t,
\end{align}
where $\mathbf{F}_t$ denotes the state transition matrix that defines the dynamics of the state vector, $\mathbf{G}_t$ represents the control-input matrix modulating the influence of the process noise, and $\mathbf{w}_t$ is assumed to follow a Gaussian distribution with zero mean and covariance matrix $\mathbf{Q}_t$.

The observation model relates the observed data $\mathbf{y}_t$ to the state vector through:
\begin{align}
\mathbf{y}_t = \mathbf{H}_t \mathbf{x}_t + \mathbf{v}_t,
\end{align}
where $\mathbf{H}_t$ is the observation matrix facilitating the mapping from the state space to the observed data, and $\mathbf{v}_t$ is the observation noise, typically modeled as Gaussian with zero mean and covariance matrix $\mathbf{R}_t$.

The efficacy of state space models in capturing the dynamics of various systems hinges on the precise characterization of the matrices $\mathbf{F}_t$, $\mathbf{G}_t$, $\mathbf{H}_t$, and the noise processes $\mathbf{Q}_t$ and $\mathbf{R}_t$. These matrices may be static or may vary with time, reflecting the changing dynamics of the system under study. The estimation of the latent states $\mathbf{x}_t$ from the observations $\mathbf{y}_t$ generally employs recursive algorithms such as the Kalman filter for linear models and particle filters for nonlinear variants. These methodologies rely on assumptions regarding the initial state distribution and the statistical properties of the noise components. This foundational description underscores the adaptability of state space models in addressing a myriad of applications across fields, where they are pivotal in modeling dynamic systems subjected to stochastic disturbances and observational noise.

\section{MemoryMamba}
In this section, we first elaborate on the overall architecture of MemoryMamba. Subsequently, we introduce our Mem-SSM Block, which includes our proposed Coarse- and Fine-Grained Memory Encoding and the Fusion Module.

\subsection{Overall Architecture}
\begin{figure}[t]
    \centering
    \includegraphics[width=1\linewidth]{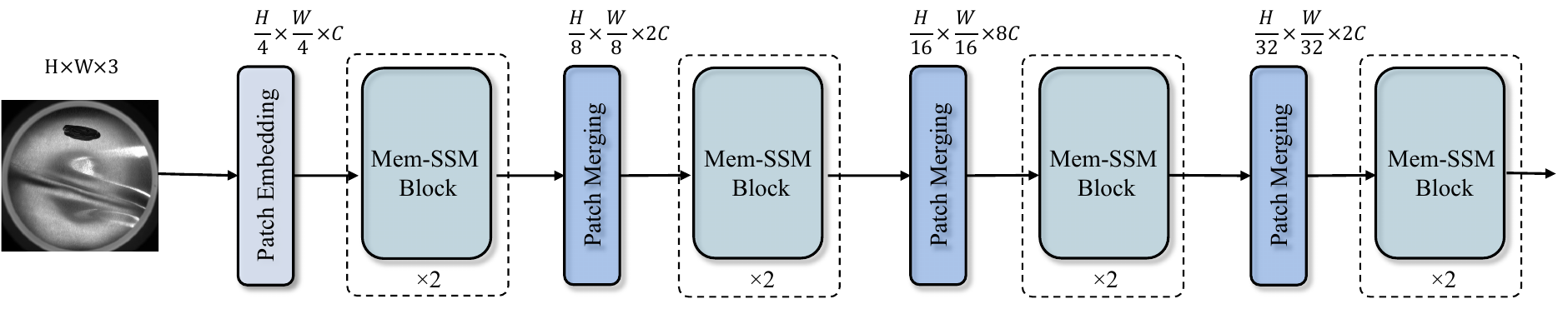}
    \caption{Overview of our method.}
    \label{fig:model}
\end{figure}

Given an image $\mI$ with size of $H \times W \times 3$, the MemoryMamba model starts with the Patch Embedding procedure, as shown in Figure~\ref{fig:model}. The input image $\mI$ is transformed into embedded patch features $\mF_0$ with dimensions of $\frac{H}{4} \times \frac{W}{4} \times C$, i.e.,
\begin{align}
\mF_0 = \PatchEmbed(\mI), \label{eq:patch_embed}
\end{align}
where $\mF_0$ $\in$ $\R^{\frac{H}{4} \times \frac{W}{4} \times C}$ denotes the embedded patch features. After the Patch Embedding, we employ the Mem-SSM Blocks to iteratively refine the feature representations, i.e.,
\begin{align}
\mF_i = \MemSSMBlock(\mF_{i-1}), \label{eq:mem_ssm_block}
\end{align}
where $\mF_i$ and $\mF_{i-1}$ are the feature sets at the input and output of the $i$-{th} Mem-SSM Block, respectively. Our model consists of $N$ Mem-SSM Blocks. Each block operation further compresses spatial dimensions and augments the channel capacity, effectively trading spatial granularity for feature depth. The final output $\mF_{N}$ is a high-dimensional representation that encapsulates the image's semantic content.

The final output $\mF_{N}$ is then fed into a Multilayer perceptron ($\MLP$) for classification, i.e.,
\begin{align}
\vh = \MLP(\mF_{N}), \label{eq:classifier_head}
\end{align}
where $\vh$ denotes the hidden vector of the $\MLP$.
Then, we utilize the softmax function to compute the predicted probability distribution over the classes, i.e.,
\begin{align}
\vp = \softmax(\vh), \label{eq:softmax}
\end{align}
where $\vp$ $\in$ $\R^{K}$ denotes the predicted probability distribution over $K$ classes.
Finally, our method can be trained end-to-end by minimizing the cross-entropy loss between the predicted probability distribution $\vp$ and the ground-truth label $y$,i.e.,
\begin{align}
\gL_{cls} = -\vy \cdot \log(\vp), \label{eq:loss}
\end{align}
where $\gL_{cls}$ denotes the cross-entropy loss.

\begin{figure}[t]
    \centering
    \includegraphics[width=1\linewidth]{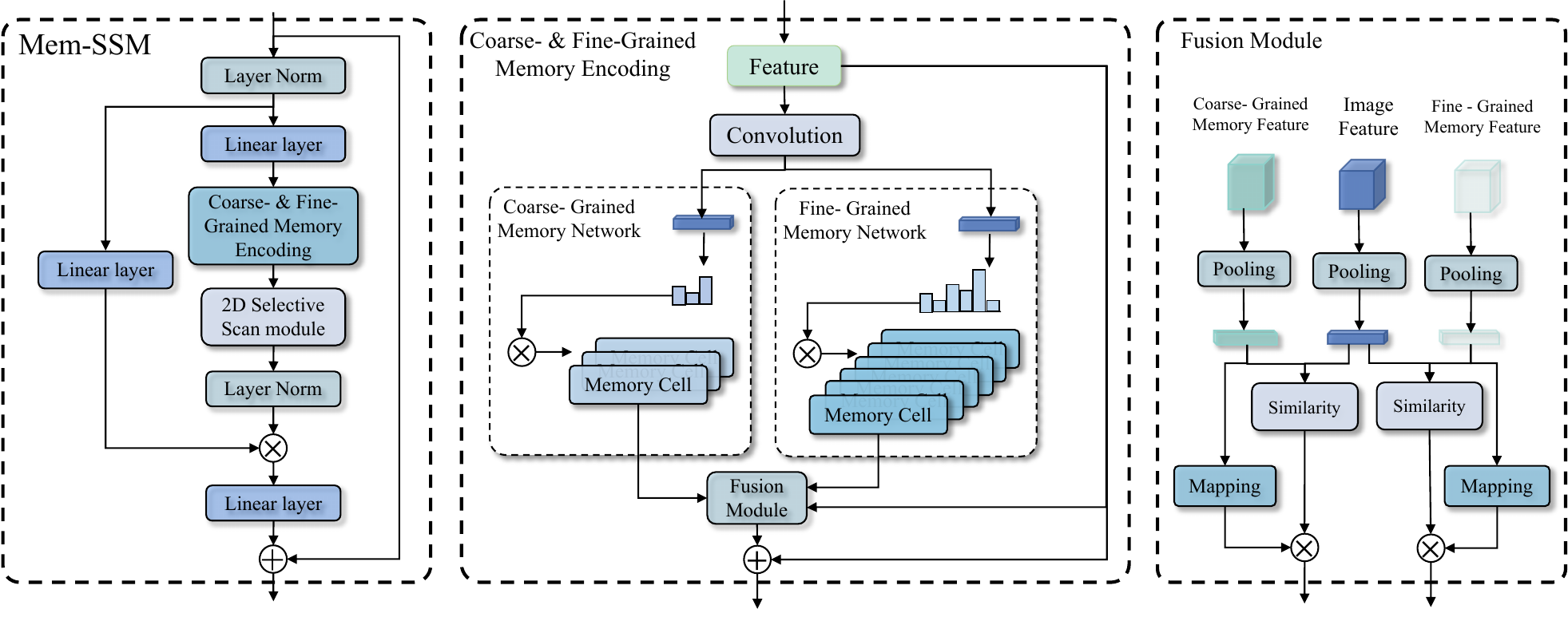}
    \caption{Details of Mem-SSM Block.}
    \label{fig:module}
\end{figure}

\subsection{Mem-SSM Block}
Due to the limited availability of defect samples, we design the Mem-SSM Block, which integrates memory networks to learn memory information from training samples. As shown in Figure~\ref{fig:module}, the Mem-SSM Block consists of memory encoding and selective scanning, thereby extracting a robust representation of the input that is particularly sensitive to the nuances required for accurate defect detection.

The block operates on the input features $\mF_{i-1}$ by initially passing them through a linear layer to produce an intermediate feature set $\mZ_i$, which can be formally expressed as:
\begin{align}
\mZ_i = \Linear(\mF_{i-1}; \theta_{\Linear}), \label{eq:linear}
\end{align}
where $\theta_{\Linear}$ represents the learnable parameters of the linear transformation.

Following the initial linear transformation, the Mem-SSM Block introduces the Coarse- and Fine-Grained Memory Encoding module, which is responsible for capturing and encoding hierarchical memory states. The memory encoding can be formulated as:
\begin{align}
\mM_i^{\text{coarse}}, \mM_i^{\text{fine}} = \MemoryEncode(\mZ_i; \Theta_{\MemoryEncode}), \label{eq:memory_encode}
\end{align}
where $\mM_i^{\text{coarse}}$ and $\mM_i^{\text{fine}}$ denote the coarse and fine memory states, respectively, and $\Theta_{\MemoryEncode}$ denotes the set of parameters governing the memory encoding process.

To integrate these encoded memory states with the intermediate features, a fusion module is introduced:
\begin{align}
\bar{\mF_i} = \FusionModule(\mM_i^{\text{coarse}}, \mM_i^{\text{fine}}, \mZ_i; \Theta_{\Fusion}), \label{eq:fusion}
\end{align}
where $\bar{\mF_i}$ denotes the fused feature set and $\Theta_{\Fusion}$ represents the parameters of the fusion module.

The fused features $\bar{\mF_i}$ are then processed through the 2D Selective Scan module, which selectively emphasizes important feature responses while suppressing less relevant ones:
\begin{align}
\mS_i = \SelectiveScan(\bar{\mF_i}; \theta_{\SelectiveScan}), \label{eq:selective_scan}
\end{align}
where $\mS_i$ is the selectively scanned feature set, and $\theta_{\SelectiveScan}$ are the parameters of this module.

The output of the Selective Scan module is then normalized using a Layer Normalization step:
\begin{align}
\mN_i = \LayerNorm(\mS_i; \gamma, \beta), \label{eq:layer_norm}
\end{align}
where $\gamma$ and $\beta$ are the parameters for scaling and shifting during the normalization process.

The normalized features are combined with the original input features through a residual connection, followed by a second linear transformation:
\begin{align}
\mF_i = \Linear(\mF_{i-1} + \mN_i; \theta_{\Linear}^{\prime}), \label{eq:linear_residual}
\end{align}
where $\theta_{\Linear}^{\prime}$ represents the learnable parameters of the second linear layer, which completes the processing within the Mem-SSM Block. The residual connection helps preserve the original feature information while allowing the network to learn modifications to the feature set adaptively.

By repeatedly applying the Mem-SSM Block, MemoryMamba progressively refines the visual representation.

\subsubsection{Coarse- \& Fine-Grained Memory Encoding}
The Coarse- \& Fine-Grained Memory Encoding is the crux of our Mem-SSM Block, as it underpins the model's ability to discern and encode varying levels of feature details. This encoding mechanism is pivotal for nuanced tasks such as defect detection, where fine distinctions can determine the correct classification.

The memory encoding process begins with the application of a convolution operation to the intermediate feature set $\mZ_i$, which serves to extract spatial hierarchies within the data:
\begin{align}
\Tilde{\mF_i} = \Convolution(\mZ_i; \theta_{\Convolution}), \label{eq:convolution}
\end{align}
where $\Tilde{\mF_i}$ is the convolution-processed feature set and $\theta_{\Convolution}$ denotes the convolutional layer's parameters.

Upon obtaining $\Tilde{\mF_i}$, we proceed to map these features into memory vectors that embody the memory state at both coarse and fine levels. This mapping is achieved through a series of transformations that are designed to preserve the spatial correlations within the feature maps while reducing dimensionality to the desired memory size:
\begin{align}
\vh_c &= {\MLP}_{c}(\Tilde{\mF_i}),  \vh_c \in \R^{c}, \notag\\
\vh_f &= {\MLP}_{f}(\Tilde{\mF_i}),  \vh_f \in \R^{f}, \label{eq:memory_size}
\end{align}
where $\vh_c$ and $\vh_f$ represents the memory query vectors, and $c$ and $f$ are sizes of coarse- \& fine-grained memory network.

Upon obtaining $\Tilde{\mF_i}$, we proceed to map these features into memory vectors that embody the memory state at both coarse and fine levels. This mapping is achieved through a series of transformations that are designed to preserve the spatial correlations within the feature maps while reducing dimensionality to the desired memory size:
\begin{align}
\vh_c &= {\MLP}_{c}(\Tilde{\mF_i}),  \vh_c \in \R^{c}, \notag \\
\vh_f &= {\MLP}_{f}(\Tilde{\mF_i}),  \vh_f \in \R^{f}, \label{eq:memory_feature}
\end{align}
where $\vh_c$ and $\vh_f$ represents the memory query vectors, and $c$ and $f$ are sizes of coarse- \& fine-grained memory network.

To assign relevance to these memory query vectors, a softmax layer is applied to generate a set of attention weights, thereby enabling the model to focus on the most pertinent memory vectors during the retrieval process:
\begin{align}
\alpha_c &= \softmax(\vh_c) \notag\\
\alpha_f &= \softmax(\vh_f)
\end{align}
where $\alpha_c$ and $\alpha_f$ represent the attention weights for memory networks.

The final step in memory encoding involves the aggregation of memory knowledge by weighting the raw memory vectors with the attention weights, which is formalized as follows:
\begin{align}
\bar{\mM_c} &= \sum_{j}\alpha_{j} \cdot \mM_{c,j}, \notag \\
\bar{\mM_f} &= \sum_{t}\alpha_{t} \cdot \mM_{f,t}, \label{eq:memory_retrieval}
\end{align}
where $\bar{\mM_c}$ and $\bar{\mM_f}$ symbolize the aggregated memory vectors at both coarse and fine levels, and $j$ and $t$ index the individual memory vectors. The memory vectors $\bar{\mM_c}$ and $\bar{\mM_f}$ are subsequently made available to the Fusion Module, where they are fused with $\mZ_i$.

\subsubsection{Fusion Module}
The Fusion Module is designed to merge the information from both the Coarse- and Fine-Grained Memory Encoding with the intermediate feature set $\mZ_i$. The fusion process begins with the alignment of the memory vectors with the intermediate features, i.e.,
\begin{align}
\vv_c &= \Pooling(\bar{\mM_c}),  \notag\\
\vv_f &= \Pooling(\bar{\mM_f}), \\
\vz_i &= \Pooling(\mZ_i), 
\end{align}
Then, we calculate the similarity between memory vectors and the feature:
\begin{align}
\beta_c &= \similarity(\vv_c, \vz_i) \notag\\
\beta_f &= \similarity(\vv_f, \vz_i)
\end{align}
where $\beta_c$ and $\beta_f$ are the similarity scores between the coarse and fine memory vectors and the intermediate feature set respectively.

The similarity scores are used to modulate the contribution of the memory features. This is achieved through a weighting mechanism, which amplifies features that are relevant and suppresses the less significant ones:
\begin{align}
\vw_c &= \beta_c \cdot \vv_c,  \notag\\
\vw_f &= \beta_f \cdot \vv_f,
\end{align}
where $\vw_c$ and $\vw_f$ represent the weighted coarse and fine memory vectors. Then, we extend \(\vw_c\) and \(\vw_f\) to match the dimensionality of \(\mZ_i\). These expanded vectors are then added to \(\mZ_i\) to form the enhanced feature set \(\bar{\mF_i}\). 

\subsection{Memory Network Optimization}
To enhance the performance of the MemoryMamba architecture, specialized optimization strategies are employed for the Coarse-Grained and Fine-Grained Memory Networks. These strategies are designed to refine the memory encoding processes by leveraging both classification loss and unique memory-based losses. 

\subsubsection{Coarse-Grained Memory Network Optimization}
The Coarse-Grained Memory Network is optimized using a contrastive learning approach, which leverages the query memory vectors $\vh_c$ from different classes. This method encourages the network to distinguish between the coarse features of various classes more effectively, i.e.,
\begin{align}
\gL_{\contrastive} = \sum_{k=1}^{K} \sum_{\substack{j=1 \\ j \neq k}}^{K} \max(0, \delta - \cos(\vh_{c,k}, \vh_{c,j})),
\end{align}
where $\vh_{c,k}$ and $\vh_{c,j}$ are the query memory vectors of the $k$-th and $j$-th class, respectively, $\cos$ denotes the cosine similarity, and $\delta$ is a margin that defines the minimum acceptable distance between classes. 

\subsubsection{Fine-Grained Memory Network Optimization}
For the Fine-Grained Memory Network, optimization is focused on maximizing the mutual information between the intermediate features $\mZ_i$ and their corresponding memory representations $\bar{\mM_f}$, which have been processed through an MLP. The mutual information is maximized to ensure that the memory network captures detailed and relevant features that are crucial for fine-grained tasks, i.e.,
\begin{align}
\gL_{\text{NCE}} = -\mathbb{E}_{(\mZ_i, \bar{\mM_f})} \left[\log \frac{e^{\text{sim}(\mZ_i, \bar{\mM_f})}}{\sum_{\hat{\mM}_f \in \mathcal{N}} e^{\text{sim}(\mZ_i, \hat{\mM}_f)}} \right],
\end{align}
where $\text{sim}(\cdot, \cdot)$ denotes a similarity metric (e.g., dot product), and $\mathcal{N}$ represents a set of negative samples drawn from the memory that are not corresponding to $\mZ_i$. 

\subsection{Overall Training Objective}
The overall training objective combines the classification loss with the memory-specific losses to train the MemoryMamba model effectively:
\begin{align}
\gL_{\text{total}} = \gL_{\text{cls}} + \lambda_c \gL_{\text{contrastive}} + \lambda_f \gL_{\text{NCE}},
\end{align}
where $\lambda_c$ and $\lambda_f$ are weighting factors that balance the contribution of the contrastive and noise-contrastive estimation losses, respectively.

\section{Experiments}
\subsection{Dataset}
\begin{wraptable}{r}{0.5\textwidth}
\vspace{-0.7cm}
\centering
\caption{Details of Defect Recognition Datasets.}
\label{tab:dataset}
\begin{tabular}{lcrr}
\toprule
\textbf{Dataset} & \textbf{Category} & \textbf{Train} & \textbf{Test} \\
\midrule
Aluminum  & 4        & 277   & 123  \\
GC10      & 10       & 1,834  & 458  \\
MT        & 6        & 1,878  & 810  \\
NEU       & 6        & 228   & 3,372 \\
\bottomrule
\end{tabular}
\end{wraptable}
The Aluminum\footnote{\url{https://aistudio.baidu.com/datasetdetail/133083}}, GC10~\cite{GC10}, MT~\cite{MT}, and NEU~\cite{NEU} datasets are crucial for evaluating the performance of the defect recognition models. Each dataset includes a distinct number of categories and a split between training and testing data, as shown in Table~\ref{tab:dataset}. 
To assess model performance, we calculate the following metrics:
Accuracy (ACC), Precision (Prec), Recall (Rec), and F1 Score.
The comprehensive evaluation using these metrics allows us to thoroughly assess the performance of our defect recognition models, ensuring that they are robust and effective across different scenarios and dataset characteristics.

\subsection{Experimental Setting}
In this study, we utilized the Adam optimizer  \cite{DBLP:journals/corr/KingmaB14} to facilitate the learning process of our model. The learning rate is set to $2 \times 10^{-5}$, and the training process is 10 epochs. The weight decay was implemented at a rate of 0.01 to regularize and prevent the co-adaptation of neurons. Additionally, we incorporated a linear learning rate decay over the course of training, and a warm-up phase, accounting for 5\% of the total training duration, was also integrated. During this phase, the learning rate gradually increased from zero to the set initial rate. The batch size is set at 64 to optimize the model. Training is conducted using an NVIDIA 80 GB A100 GPU. The comparison methods include ResNet~\cite{Resnet}, DeiT~\cite{DeiT}, Swin-Transformer (Swin~\cite{Swin}), and Vmamba~\cite{liu2024vmamba}.

\subsection{Results}
\begin{table}[t]\small
\centering
\begin{minipage}{.48\linewidth}
\centering
\caption{Performance Comparison on Aluminum Dataset.}
\label{tab:Aluminum_comparison}
\begin{tabular}{lcccc}
\toprule
Method & ACC & Prec & Rec & F1 \\
\midrule
ResNet18 & 0.36 & 0.36 & 0.38 & 0.37 \\
ResNet50 & 0.42 & 0.42 & 0.39 & 0.40 \\
ResNet101 & 0.51 & 0.51 & 0.41 & 0.45 \\
ResNet152 & 0.59 & 0.59 & 0.52 & 0.55 \\
DeiT-S & 0.36 & 0.36 & 0.28 & 0.32 \\
DeiT-B & 0.52 & 0.52 & 0.45 & 0.48 \\
Swin-T & 0.24 & 0.24 & 0.35 & 0.28 \\
Swin-S & 0.49 & 0.49 & 0.51 & 0.50 \\
Swin-B & 0.62 & 0.62 & 0.55 & 0.58 \\
Vmamba-T & 0.35 & 0.36 & 0.38 & 0.37 \\
Vmamba-S & 0.55 & 0.55 & 0.59 & 0.57 \\
Vmamba-B & 0.65 & 0.65 & 0.52 & 0.58 \\
MemoryMamba & \bf 0.75 & \bf 0.75 & \bf 0.54 & \bf 0.63 \\
\bottomrule
\end{tabular}
\end{minipage}
\hfill
\begin{minipage}{.48\linewidth}
\centering
\caption{Performance Comparison on GC10 Dataset.}
\label{tab:GC10_comparison}
\begin{tabular}{lcccc}
\toprule
Method          & ACC & Prec & Rec & F1   \\ 
\midrule
ResNet18        & 0.55     & 0.55      & 0.53   & 0.48 \\
ResNet50        & 0.71     & 0.71      & 0.67   & 0.67 \\
ResNet101       & 0.74     & 0.74      & 0.68   & 0.66 \\
ResNet152       & 0.77     & 0.77      & 0.73   & 0.71 \\
DeiT-S          & 0.83     & 0.83      & 0.79   & 0.79 \\
DeiT-B          & 0.84     & 0.84      & 0.82   & 0.82 \\
Swin-T          & 0.71     & 0.71      & 0.75   & 0.70 \\
Swin-S          & 0.79     & 0.79      & 0.70   & 0.70 \\
Swin-B          & 0.86     & 0.86      & 0.78   & 0.79 \\
Vmamba-T        & 0.79     & 0.79      & 0.80   & 0.78 \\
Vmamba-S        & 0.83     & 0.83      & 0.78   & 0.79 \\
Vmamba-B        & 0.83     & 0.83      & 0.84   & 0.82 \\
MemoryMamba     & \bf 0.90     & \bf 0.90      & \bf 0.89   & \bf 0.89 \\
\bottomrule
\end{tabular}
\end{minipage}
\end{table}
Our experimental results across four datasets, i.e., Aluminum, GC10, MT, and NEU, demonstrate the superior performance of the MemoryMamba model. The comparison results are shown in Table~\ref{tab:Aluminum_comparison}, Table~\ref{tab:GC10_comparison}, Table~\ref{tab:MT_comparison}, andTable~\ref{tab:NEU_comparison}.In comparison to traditional architectures like ResNet and transformer-based models such as DeiT and Swin Transformers, MemoryMamba consistently achieved the highest scores in Accuracy, Precision, Recall, and F1 Score. Notably, it outperformed in challenging defect detection scenarios, achieving as high as 99\% in all evaluated metrics on the NEU dataset. The integration of Coarse- and Fine-Grained Memory Encoding significantly enhances the model's ability to capture detailed contextual information, thus improving its effectiveness in complex visual pattern recognition tasks across diverse conditions. 

\begin{table}[t]\small
\centering
\begin{minipage}{.48\linewidth}
\centering
\caption{Performance Comparison on MT Dataset}
\label{tab:MT_comparison}
\begin{tabular}{lcccc}
\toprule
Method         & ACC & Prec & Rec & F1   \\ \midrule
ResNet18       & 0.69     & 0.69      & 0.56   & 0.55 \\
ResNet50       & 0.83     & 0.83      & 0.71   & 0.72 \\
ResNet101      & 0.86     & 0.86      & 0.77   & 0.79 \\
ResNet152      & 0.88     & 0.88      & 0.73   & 0.78 \\
DeiT-S         & 0.80     & 0.80      & 0.75   & 0.75 \\
DeiT-B         & 0.84     & 0.84      & 0.77   & 0.78 \\
Swin-T         & 0.82     & 0.82      & 0.62   & 0.67 \\
Swin-S         & 0.83     & 0.83      & 0.71   & 0.74 \\
Swin-B         & 0.88     & 0.88      & 0.76   & 0.80 \\
Vmamba-T       & 0.87     & 0.87      & 0.89   & 0.86 \\
Vmamba-S       & 0.91     & 0.91      & 0.91   & 0.90 \\
Vmamba-B       & 0.92     & 0.92      & 0.87   & 0.89 \\
MemoryMamba    & \bf 0.96     & \bf 0.96      & \bf 0.97   & \bf 0.96 \\ \bottomrule
\end{tabular}
\end{minipage}
\hfill
\begin{minipage}{.48\linewidth}
\centering
\caption{Model performance comparison on NEU Dataset.}
\label{tab:NEU_comparison}
\begin{tabular}{lcccc}
\toprule
Method         & ACC & Prec & Rec & F1   \\ 
\midrule
ResNet18       & 0.86     & 0.86      & 0.78   & 0.72 \\ 
ResNet50       & 0.89     & 0.86      & 0.89   & 0.87 \\ 
ResNet101      & 0.94     & 0.94      & 0.93   & 0.93 \\ 
ResNet152      & 0.94     & 0.94      & 0.94   & 0.94 \\ 
DeiT-S         & 0.89     & 0.89      & 0.86   & 0.86 \\ 
DeiT-B         & 0.93     & 0.93      & 0.91   & 0.91 \\ 
Swin-T         & 0.88     & 0.87      & 0.88   & 0.88 \\ 
Swin-S         & 0.91     & 0.91      & 0.85   & 0.84 \\ 
Swin-B         & 0.92     & 0.92      & 0.92   & 0.92 \\ 
Vmamba-T       & 0.91     & 0.92      & 0.89   & 0.90 \\ 
Vmamba-S       & 0.92     & 0.91      & 0.91   & 0.91 \\ 
Vmamba-B       & 0.94     & 0.94      & 0.94   & 0.94 \\ 
MemoryMamba    & \bf 0.99     & \bf 0.99      & \bf 0.99   & \bf 0.99 \\
\bottomrule
\end{tabular}
\end{minipage}
\end{table}

\subsection{Ablation Study}
\begin{table*}[t]\small
\centering
\caption{Ablation Study of MemoryMamba.}
\begin{tabular}{lcccccccc}
\toprule
\multirow{2}{*}{Method} & \multicolumn{2}{c}{Aluminum} & \multicolumn{2}{c}{GC10} & \multicolumn{2}{c}{MT} & \multicolumn{2}{c}{NEU} \\ 
\cmidrule(r){2-3} \cmidrule(r){4-5} \cmidrule(r){6-7} \cmidrule(r){8-9}
       & ACC & F1 & ACC & F1 & ACC & F1 & ACC & F1 \\ 
\midrule
MemoryMamba                  & \bf 0.75 & \bf 0.63 & \bf 0.90 & \bf 0.89 & \bf 0.96 & \bf 0.96 & \bf 0.99 & \bf 0.99 \\\midrule
$\diamondsuit$ w/o CMN                      & 0.71 & 0.61 & 0.87 & 0.86 & 0.95 & 0.94 & 0.97 & 0.97 \\
$\diamondsuit$ w/o FMN                      & 0.72 & 0.62 & 0.88 & 0.87 & 0.95 & 0.94 & 0.98 & 0.98 \\
$\diamondsuit$ w/o Fusion                   & 0.72 & 0.61 & 0.88 & 0.88 & 0.95 & 0.94 & 0.97 & 0.98 \\
$\diamondsuit$ w/o CMN, Fusion              & 0.70 & 0.60 & 0.86 & 0.85 & 0.94 & 0.92 & 0.96 & 0.96 \\
$\diamondsuit$ w/o FMN, Fusion              & 0.70 & 0.60 & 0.87 & 0.85 & 0.94 & 0.93 & 0.96 & 0.96 \\
$\diamondsuit$ w/o CMN, FMN, Fusion         & 0.65 & 0.58 & 0.83 & 0.82 & 0.92 & 0.89 & 0.94 & 0.94 \\
\bottomrule
\end{tabular}
\end{table*}
In the ablation study, we evaluate the contribution of the Coarse-Grained Memory Network (CMN), Fine-Grained Memory Network (FMN), and the Fusion Module within the MemoryMamba architecture. The removal of each component consistently led to a decrease in both accuracy and F1 scores, underscoring their individual and collective importance. The most significant performance drops occurred when multiple components were excluded simultaneously, highlighting their synergistic effect. These findings emphasize the critical roles of CMN and FMN in capturing hierarchical feature details and the Fusion Module in effectively integrating these features, which are vital for the model’s performance and robustness in industrial defect detection tasks.

\subsection{Impact on Fusion Module}
\begin{figure}[t]
    \centering
    \includegraphics[width=0.48\linewidth]{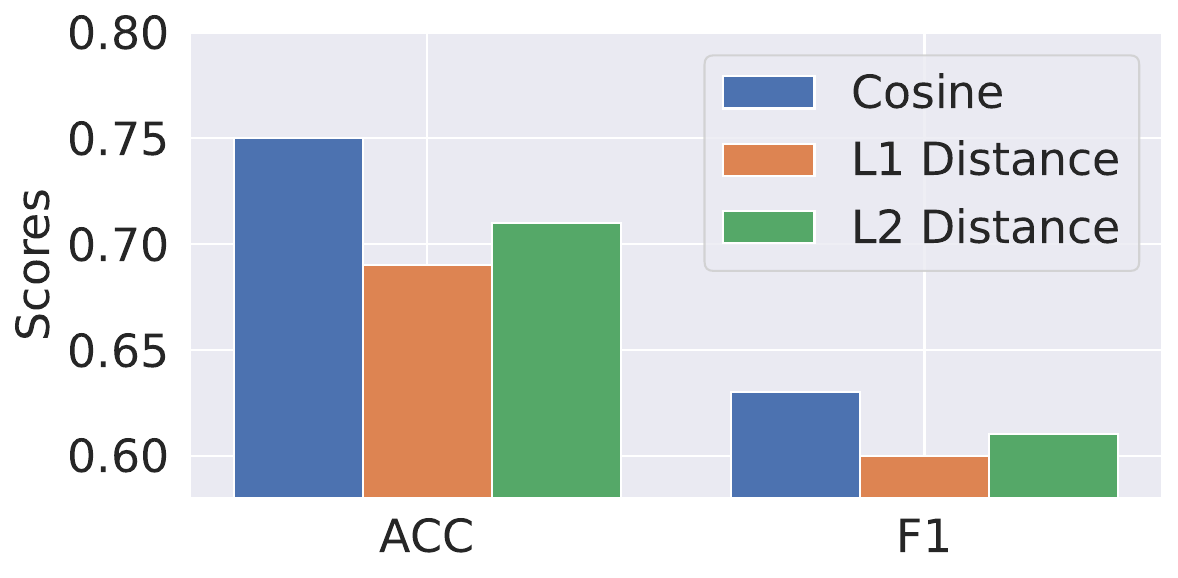}
    \hfill
    \includegraphics[width=0.48\linewidth]{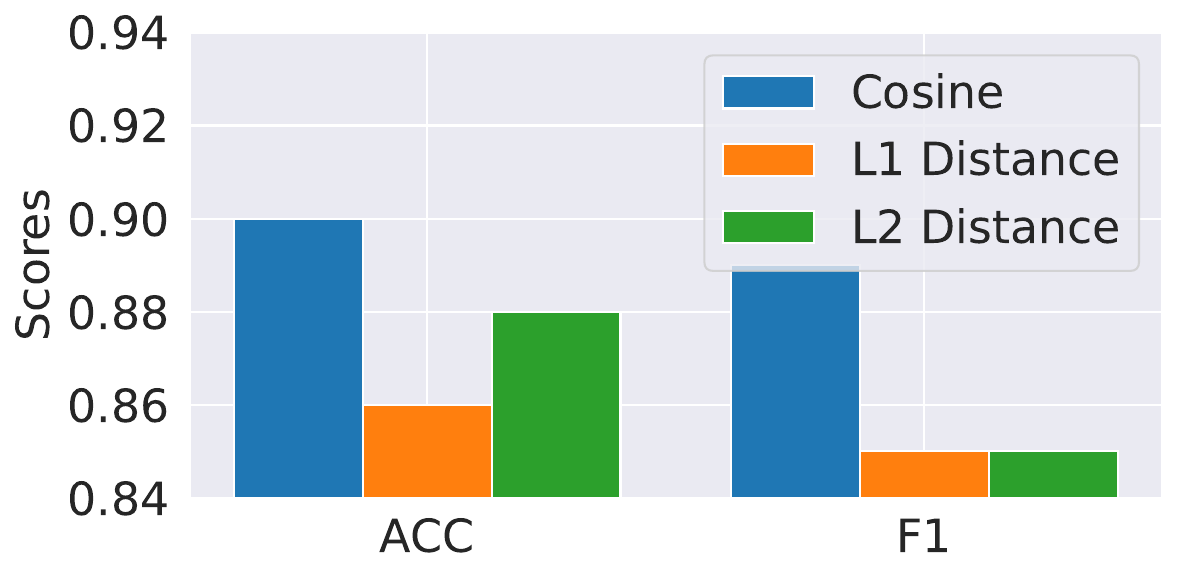}
    \caption{Comparison of similarity evaluation methods for Fusion Module on Aluminum (left) and GC10 (right) datasets.}
    \label{fig:similarity}
\end{figure}
The Fusion Module is critical to MemoryMamba, directly influencing its classification accuracy and F1 scores, as demonstrated in Figure \ref{fig:similarity}. This module's role in integrating Coarse- and Fine-Grained Memory Encodings with the feature set is validated through performance metrics on Aluminum and GC10 datasets using different similarity evaluation methods. Our findings highlight that Cosine similarity achieves higher performance on Aluminum and GC10. The choice of similarity metric thus plays a crucial role in tuning the Fusion Module for optimal defect detection performance.

\subsection{Impact on Memory Networks}
\begin{figure}[!t]
    \centering
    \includegraphics[width=0.48\linewidth]{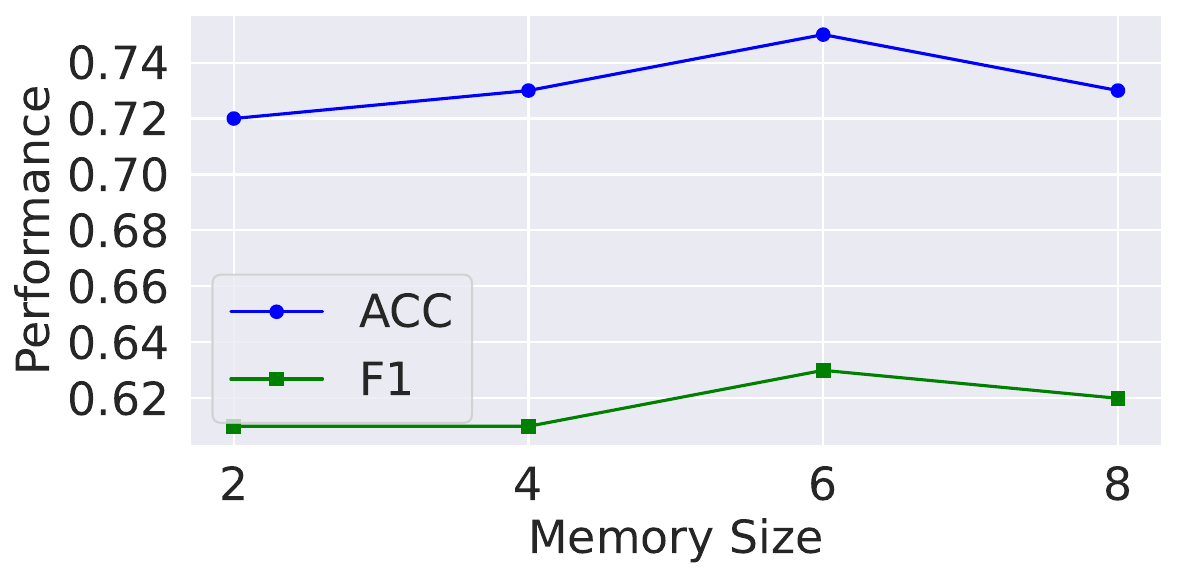}
    \hfill
    \includegraphics[width=0.48\linewidth]{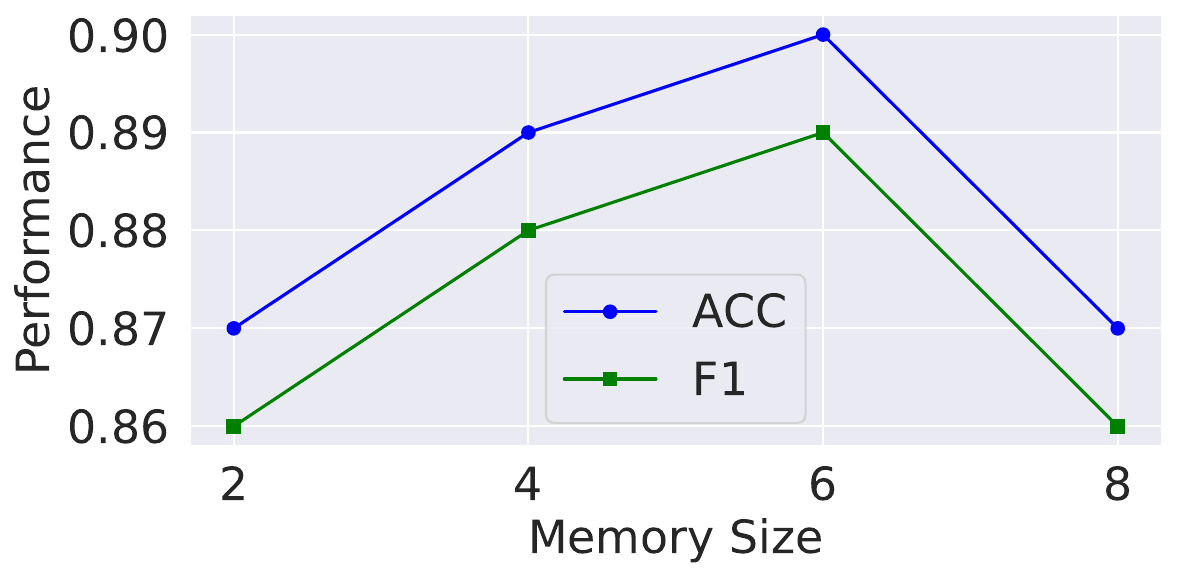}
    \caption{Investigation of different memory size for Coarse-grained Memory network on Aluminum (left) and GC10 (right) datasets.}
    \label{fig:global_memory}
\end{figure}
\begin{figure}[!t]
    \centering
    \includegraphics[width=0.48\linewidth]{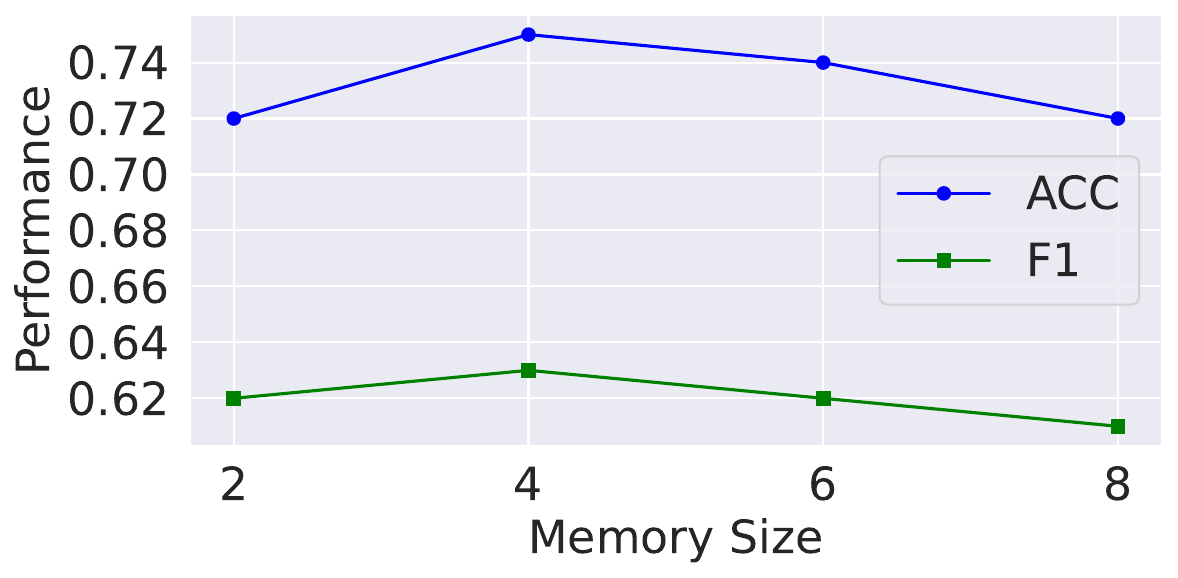}
    \hfill
    \includegraphics[width=0.48\linewidth]{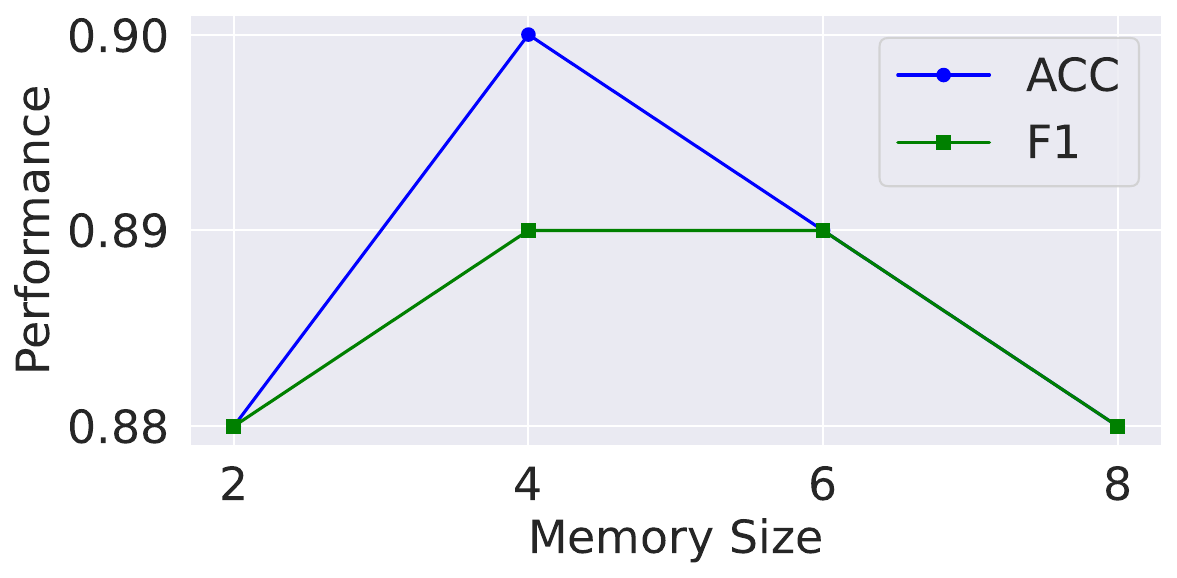}
    \caption{Investigation of different memory size for Fine-grained Memory network on Aluminum (left) and GC10 (right) datasets.}
    \label{fig:local_memory}
\end{figure}
The performance of memory networks is crucial for the robustness of defect recognition systems. Our investigation, illustrated in Figures \ref{fig:global_memory} and \ref{fig:local_memory}, highlights the influence of memory size on the accuracy (ACC) and F1 Score of Coarse-grained and Fine-grained Memory networks, respectively. For the Aluminum dataset, the Coarse-grained Memory network exhibits a peak performance at a memory size of 4, with diminishing returns as the size increases. 
On the more complex GC10 dataset, both memory network types show a significant performance variation with memory size changes. 
These observations suggest that the optimal memory size is contextually dependent on the dataset granularity and the network's memory type.

\subsection{Similarity Calculation on Memory Networks}
\begin{figure}[!t]
    \centering
    \includegraphics[width=0.48\linewidth]{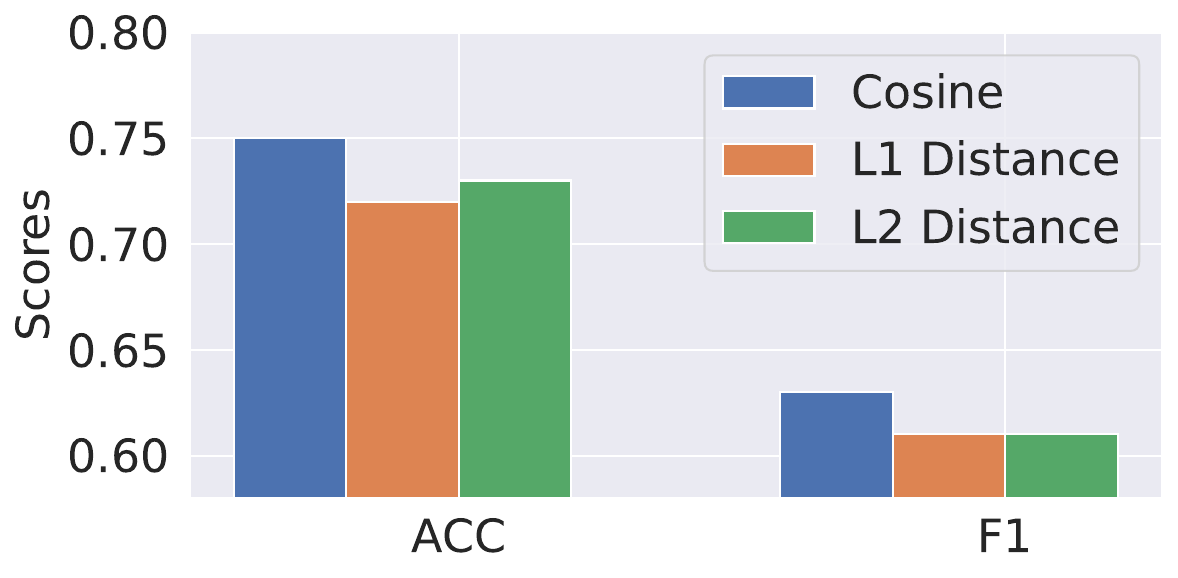}
    \hfill
    \includegraphics[width=0.48\linewidth]{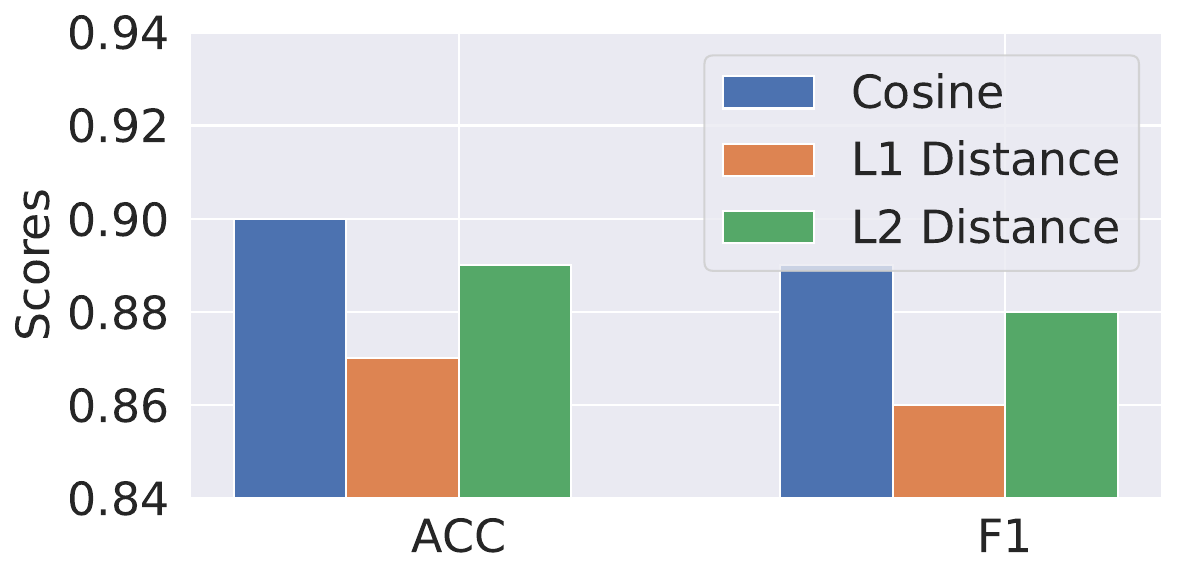}
    \caption{Impact of different similarity calculation methods for Coarse-grained Memory network on Aluminum (left) and GC10 (right) datasets.}
    \label{fig:global_similarity}
\end{figure}
\begin{figure}[!t]
    \centering
    \includegraphics[width=0.48\linewidth]{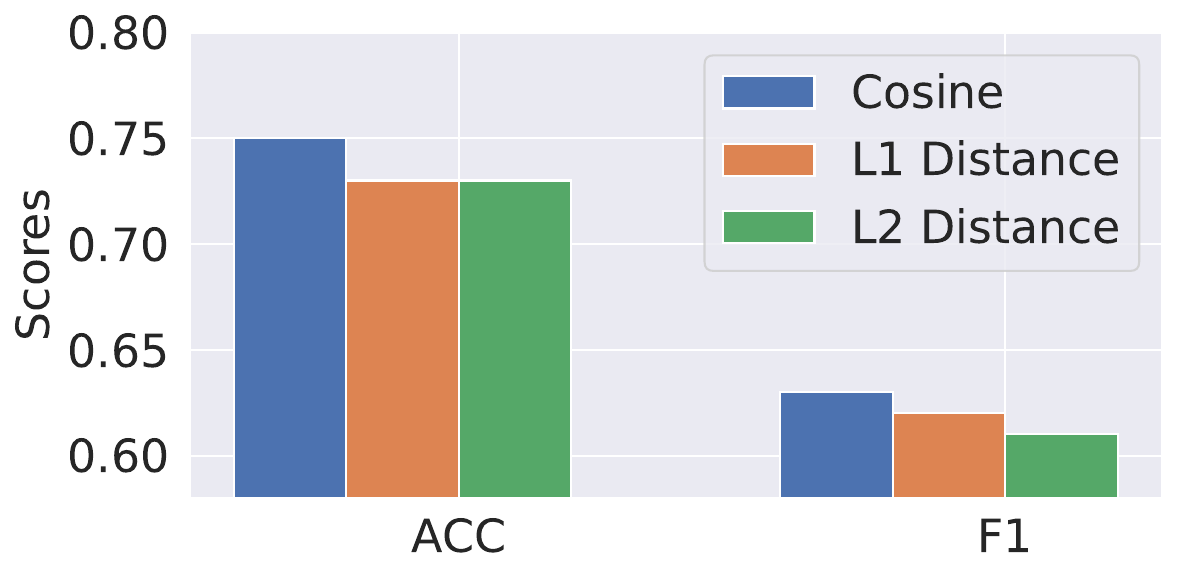}
    \hfill
    \includegraphics[width=0.48\linewidth]{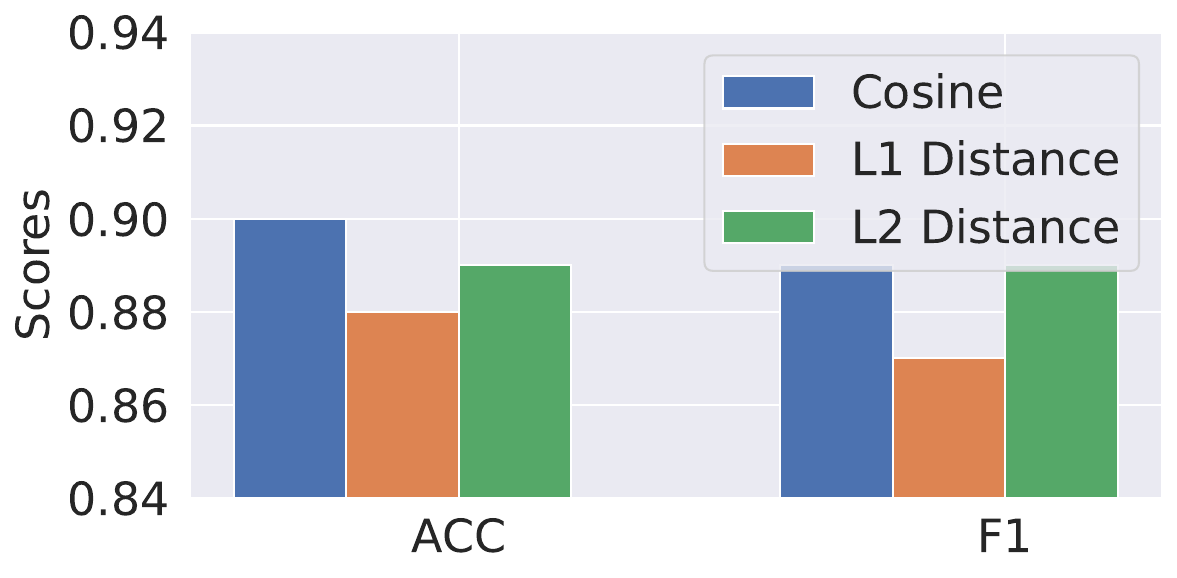}
    \caption{Impact of different similarity calculation methods for Fine-grained Memory network on Aluminum (left) and GC10 (right) datasets.}
    \label{fig:local_similarity}
\end{figure}
For memory networks, the choice of similarity calculation method can significantly influence the performance of the model. To investigate this, we employed three different similarity measures: cosine similarity, L1 distance, and L2 distance, on coarse- and fine-grained memory networks.
Figures \ref{fig:global_similarity} and \ref{fig:local_similarity} illustrate the impact of these methods on the Aluminum and GC10 datasets. Our observations reveal that cosine similarity consistently outperforms the L1 and L2 distances in the accuracy (ACC) and F1 score.

\section{Conclusion}
In this work, we integrate state space models with memory augmentation to address the limitations of other methods in defect recognition systems.
We demonstrate that MemoryMamba's unique architecture, which combines coarse-grained and fine-grained memory networks with a novel fusion module, effectively captures and utilizes historical defect-related data. This capability allows for enhanced detection of complex and subtle defects that previous models may overlook. The application of contrastive learning and mutual information maximization strategies in optimizing these memory networks further enriches the robustness and accuracy of the defect detection process.
The experimental results from four distinct industrial datasets have underscored MemoryMamba's superiority over existing technologies such as CNNs and Vision Transformers.

\bibliographystyle{plainnat}
\bibliography{ref}
\end{document}